\documentclass[sigconf]{aamas}

\usepackage{balance} % for balancing columns on the final page
\usepackage{multirow}
\doi{RIHP6974}

\makeatletter
\gdef\@copyrightpermission{
  \begin{minipage}{0.2\columnwidth}
   \href{https://creativecommons.org/licenses/by/4.0/}{\includegraphics[width=0.90\textwidth]{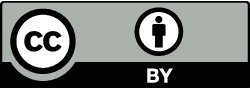}}
  \end{minipage}\hfill
  \begin{minipage}{0.8\columnwidth}
   \href{https://creativecommons.org/licenses/by/4.0/}{This work is licensed under a Creative Commons Attribution International 4.0 License.}
  \end{minipage}
  \vspace{5pt}
}
\makeatother

\setcopyright{ifaamas}
\acmConference[AAMAS '26]{Proc.\@ of the 25th International Conference
on Autonomous Agents and Multiagent Systems (AAMAS 2026)}{May 25 -- 29, 2026}
{Paphos, Cyprus}{C.~Amato, L.~Dennis, V.~Mascardi, J.~Thangarajah (eds.)}
\copyrightyear{2026}
\acmYear{2026}
\acmDOI{}
\acmPrice{}
\acmISBN{}

\acmSubmissionID{1726}

\title[AAMAS-2026 SPADE]{SPADE: Sketch-guided Path Planning Augmented with Diffusion Experts}

\author{Charbel Abi Hana}
\affiliation{
  \institution{IDEALworks GmbH}
  \city{Munich}
  \country{Germany}}
\email{charbel.abihana@idealworks.com}

\author{Tatiana Ghantous}
\affiliation{
  \institution{IMT Atlantique}
  \city{Brest}
  \country{France}}
\email{tatiana.ghantous@imt-atlantique.net}

\author{Mikael Khalil}
\affiliation{
  \institution{IMT Atlantique}
  \city{Brest}
  \country{France}}
\email{mikael.khalil@imt-atlantique.net}

\author{Anthony Rizk}
\affiliation{
  \institution{IDEALworks GmbH \& Saint Joseph University of Beirut}
  \city{Munich}
  \country{Germany}}
\email{anthony.rizk@idealworks.com}
\begin{abstract}
Path planning is essential for Autonomous Mobile Robots (AMRs). Conventional methods for incorporating human preferences into planning typically rely on either complex reward engineering or hardware-intensive solutions. Recent state-of-the-art frameworks leverage imitation learning to train behavior-specific path planning models from expert demonstrations. However, these approaches face two key limitations: limited generalization to unseen environments and low robustness in demonstration collection. To address these challenges, this work introduces an enhanced framework that focuses on two main contributions: an overhauled annotation tool built on ROS 2, and a novel training strategy that integrates diffusion-based augmentation into baseline behavioral cloning models. A dataset of expert demonstrations is provided and evaluated through ablation studies to assess the robustness of the proposed solution. The enhanced approach outperforms state-of-the-art methods with  39.1\% lower Absolute Pose Error (APE) and  33.5\% lower Fréchet Inception Distance (FID) while having 93.8\% less trainable parameters. Moreover it attains diffusion-level generalization while preserving the real-time, on-edge properties of state-of-the-art models.
\end{abstract}


\newcommand{\BibTeX}{\rm B\kern-.05em{\sc i\kern-.025em b}\kern-.08em\TeX}

\begin{document}

%%% The following commands remove the headers in your paper. For final 
%%% papers, these will be inserted during the pagination process.

\pagestyle{fancy}
\fancyhead{}

%%% The next command prints the information defined in the preamble.

\maketitle 

%%%%%%%%%%%%%%%%%%%%%%%%%%%%%%%%%%%%%%%%%%%%%%%%%%%%%%%%%%%%%%%%%%%%%%%%

\section{Introduction}
Autonomous Mobile Robots (AMRs) have become essential in modern industry, automating material handling, warehousing, and logistics operations. Their effectiveness hinges on robust path planning, the ability to navigate efficiently while avoiding obstacles in dynamic environments.
Classical planners like A* \cite{4082128} and and the Rapidly-
Exploring Random Tree (RRT) \cite{770022} excel at finding optimal routes under geometric constraints, but they struggle to encode nuanced operator preferences as simple objective functions. Real deployments show that users often want to express desired motion directly such as maintaining specific clearances, preferring certain corridors, or exhibiting smoother turns in sensitive areas. Defining such preferences introduces additional complexity to the problem, as traditional path planning algorithms struggle to accommodate user-specific intentions and requirements. These preferences often lack a formal structure that can be easily optimized using cost functions or graph-based methods. For instance, a user might favor broad, L-shaped turns when approaching docking zones—an approach that deviates from the shortest or most time-efficient route typically prioritized by standard algorithms.\\
Imitation Learning (IL) offers a compelling solution by enabling agents to learn directly from expert demonstrations. IL allows systems to replicate behaviors that reflect the desired operational subtleties \cite{10602544}. This is particularly valuable in path planning, where IL can translate the expertise of domain specialists, who may not have technical backgrounds, into intuitive demonstrations. These demonstrations help autonomous mobile robots (AMRs) adopt user-driven navigation strategies. However, IL’s effectiveness depends on the availability of high-quality demonstrations, which can be difficult to obtain in specialized environments such as industrial settings. Constraints such as limited flexibility for teleoperation and privacy concerns around RGB data collection make these demonstrations rare and costly \cite{8949040}.

End-to-end Sketch-Guided Path Planning through Imitation Learning for Autonomous Mobile Robots (SKIPP) \cite{10924509} addressed these constraints by letting operators draw desired paths on a map, converting sketches into training examples through imitation learning. SKIPP allows users to intuitively guide robots by drawing desired paths, which the system interprets and translates into executable navigation commands through an end-to-end neural architecture. Although SKIPP successfully learned the sketched behaviors, using behavioral cloning (BC) \cite{torabi2018behavioralcloningobservation} with a U-Net backbone, it faced challenges in generalizing to new, unseen maps. In this work, we address the key limitations that have hindered the scalability and generalization of SKIPP in real-world scenarios and introduce a set of architectural and training enhancements that significantly improve its robustness, adaptability and readiness for deployment. We integrate a Diffusion-guided Behavioral Cloning (DBC) pipeline, where a high-capacity diffusion model trained offline guides a compact BC network during training retaining the diffusion expert's judgment while keeping the deployable model fast and lightweight \cite{11080376}. Moreover, we improve the augmentation pipeline by training image-conditioned diffusion models and using them as experts, providing more powerful guidance signals during training. In addition, we develop an improved annotation tool built on ROS 2 to enhance data accessibility and quality for both operators and researchers, allowing more reliable and efficient demonstration collection.

The remainder of this paper is organized as follows: Section 2 provides background on imitation learning and diffusion models for path planning; Section 3 presents our proposed framework; Section 4 describes the experimental setup and evaluates the performance against baselines; Section 5 presents ablation studies examining the contribution of individual components; and Section 6 concludes the paper. The code and datasets will be made available upon publication.

\section{Background and Preliminaries}
\subsection{Imitation Learning for Path Planning}
Imitation learning enables robots to acquire skills by mimicking expert demonstrations rather than explicitly programming behaviors or defining reward functions. Among imitation learning approaches, behavioral cloning (BC) \cite{torabi2018behavioralcloningobservation} frames the learning problem as supervised learning, where a policy is trained to predict expert actions given observed states. By minimizing the discrepancy between the predicted and demonstrated actions, BC allows robots to learn complex behaviors directly from the data. However, standard BC suffers from limited generalization, particularly when encountering states not well-represented in the training distribution \cite{ross2011reductionimitationlearningstructured}, leading to compounding errors during deployment. 

Several works have explored learning custom navigation behaviors from domain expert input. \cite{doi:10.1177/0278364910369715} demonstrated learning cost functions for outdoor mobile robot navigation from demonstrated paths, while \cite{Pfeiffer_2017} used expert demonstrations to learn socially-aware navigation policies. Inverse reinforcement learning approaches \cite{10.5555/1620270.1620297} attempt to infer underlying reward functions from demonstrations but require significant computational resources. More recently, APPLD \cite{9117021} enabled operators to provide preference feedback to adapt planner parameters for context-specific behaviors. However, APPLD requires iterative human-in-the-loop feedback during deployment and assumes preferences can be captured through parameter tuning of existing planners, limiting its ability to learn fundamentally different motion patterns that may not be expressible within the base planner's parameterization.

SKIPP provided a framework enabling non-expert operators to convey desired navigation behaviors for AMRs through simple sketched demonstrations. These sketches were processed by a U-Net \cite{ronneberger2015unetconvolutionalnetworksbiomedical} trained under the behavioral cloning paradigm to generate context-specific paths. While SKIPP demonstrated promising results in learning operator preferences, the pipeline exhibited significant limitations in model generalization capabilities and the robustness of the data annotation tool.

\subsection{Diffusion for Path Planning}
Denoising Diffusion Probabilistic Models (DDPM) \cite{ho2020denoisingdiffusionprobabilisticmodels} have emerged as powerful generative models that learn to generate data by reversing a gradual noising process. DDPMs progressively add Gaussian noise to data over multiple timesteps and train a neural network to reverse this process, enabling high-quality sample generation. The forward diffusion process is defined as:
\begin{equation}
    q(x_t | x_{t-1}) = \mathcal{N}(x_t; \sqrt{1-\beta_t}x_{t-1}, \beta_t I)
\end{equation}
where $\beta_t$ controls the noise schedule. The model learns to predict the reverse process $p_\theta(x_{t-1}|x_t)$, gradually denoising samples from pure noise to generate realistic data. Subsequent work on Denoising Diffusion Implicit Models (DDIM) \cite{song2022denoisingdiffusionimplicitmodels} introduced deterministic sampling procedures that enable faster generation while maintaining sample quality.

Conditional diffusion models extend this framework by incorporating additional information to guide generation. Classifier-free guidance \cite{ho2022classifierfreediffusionguidance} emerged as an effective conditioning technique that blends conditional and unconditional predictions. Image-conditioned diffusion models \cite{saharia2022paletteimagetoimagediffusionmodels, rombach2022highresolutionimagesynthesislatent} have demonstrated remarkable success in tasks like image-to-image translation and inpainting by conditioning the denoising process on input images, enabling controlled generation based on visual context.

 Diffusion-Augmented Behavioral Cloning (DBC) \cite{chen2024diffusionmodelaugmentedbehavioralcloning} addresses the limited generalization capabilities of standard behavioral cloning by leveraging diffusion models as data augmentation mechanisms. In DBC, a high-capacity diffusion model is trained offline on expert demonstrations to capture the underlying distribution of desired behaviors. This expert model then guides the training of a compact network through augmented supervision, where the diffusion model generates additional training signals that help the compact model to generalize beyond the original demonstrations. This training guidance approach enables the deployment of lightweight models while retaining the representational power of larger diffusion models, making it particularly suitable for resource-constrained robotic systems. Our work extends this framework by introducing image-conditioned diffusion models to the DBC pipeline, enabling context-aware path generation that adapts to varying environmental configurations.

In the robotics domain, diffusion models have recently been applied to path planning and trajectory generation. Diffusion Policy \cite{chi2024diffusionpolicyvisuomotorpolicy} showed that diffusion models can effectively learn multimodal action distributions for visuomotor control. Motion Planning Diffusion (MPD) \cite{carvalho2024motionplanningdiffusionlearning} applied diffusion models to generate collision-free trajectories in continuous spaces. More recently, \cite{janner2022planningdiffusionflexiblebehavior} demonstrated that diffusion models can serve as trajectory optimizers by learning to generate feasible paths that satisfy both task objectives and physical constraints. These works highlight the potential of diffusion models to capture complex, multimodal path distributions while maintaining computational efficiency through guided sampling techniques. These works highlight the potential of diffusion models to capture complex, multimodal path distributions while maintaining computational efficiency through guided sampling techniques. However, the iterative denoising process inherent to diffusion models poses challenges for real-time robotic deployment on resource-constrained embedded systems, where inference latency and computational overhead remain critical constraints.
%%%%%%%%%%%%%%%%%%%%%%%%%%%%%%%%%%%%%%%%%%%%%%%%%%%%%%%%%%%%%%%%%%%%
\section{Proposed Method}
This work advances the SKIPP pipeline by introducing three key contributions: an overhauled annotation tool, integration of a diffusion-enhanced framework, and modeling the diffusion process as a customer-usable training tool. Specifically, we address SKIPP's most prominent limitations—expert demonstration collection and generalization—by introducing a fully re-worked data annotation tool and an enhanced training framework that integrates a diffusion model to guide the learning of smaller, deployable models, as shown in \cite{chen2024diffusionmodelaugmentedbehavioralcloning}. We extend DBC by introducing an image-conditioned diffusion model to the pipeline.
%%%%%%%%%%%%%%%%%%%%%%%%%%%%%%%%%%%%%%%%%%%%%%%%%%%%%%%%%%%%%%%%%%%%
%%%%%%%%%%%%%%%%%%%%%%%%%%%%%%%%%%%%%%%%%%%%%%%%%%%%%%%%%%%%%%%%%%%%
\subsection{Data Annotation Tool}
\label{sec:data_annotation_tool}
A primary limitation of SKIPP lies in its data acquisition mechanism. SKIPP relied on Flatsim, a 2D simulator built using NVIDIA’s ISAAC SDK \cite{Monteiro2019SimulatingRR}, where users could load robots with 2D LiDARs and simulate navigation, localization, and collision avoidance. However, the ISAAC SDK was deprecated around the time of SKIPP’s publication, severely limiting the pipeline’s maintainability and scalability. While SKIPP provided the dataset, the tool itself was not released.
In this work, we introduce a fully open-source annotation pipeline built using ROS 2, addressing these limitations and ensuring reproducibility and extensibility. We began from a 2D global occupancy grid and converted it into a 3D Gazebo world using the open-source ROS package \textit{map2gazebo} that automates 2D-to-3D map conversion.
We employed TurtleBot3, a ROS-based differential-drive mobile robot widely used for research and prototyping. Navigation was enabled through the Nav2 framework \cite{macenski2023survey}, which provides modular components for autonomous motion. Together, these modules enable real-time localization, planning, and obstacle avoidance within the simulated environment.
Using RViz2, users defined navigation paths by interactively selecting waypoints, which were published to Nav2’s action server for sequential goal execution. The resulting robot trajectories followed the user-defined shapes, allowing customizable motion generation. Data was recorded across three modalities: path, end-goal, and robot pose, extracted respectively from the \textit{/waypoints}, \textit{/amcl\_pose}, and \textit{/local\_costmap} topics.
To enhance learnability and performance, a spline-based smoothing function was applied to each path, resampling it into a continuous, evenly spaced curve. This reduced noise and irregularities, producing structured trajectories that improved training stability and generalization. Together, these components form a complete pipeline: from human-drawn sketches through trainable model encodings to navigable paths executable by AMRs.
\begin{figure}[htbp]
    \centering
        \includegraphics[width=0.9\linewidth]{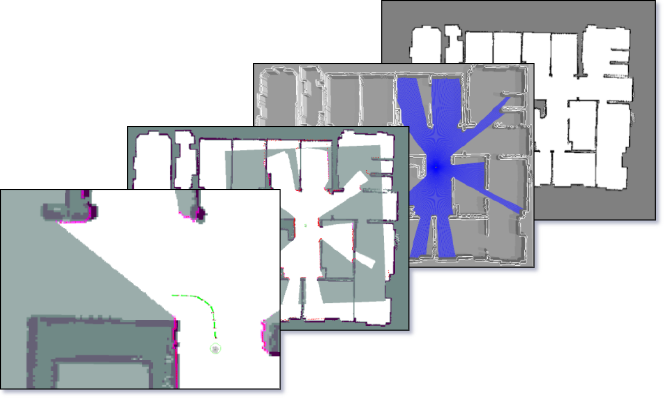}
    \caption{Global occupancy grid (rightmost), robot localization in 3D environment(second image), local occupancy grid boundaries and robot localization (third image) and finally path execution (fourth image).}
    \label{fig:annotation_tool}
\end{figure}
%%%%%%%%%%%%%%%%%%%%%%%%%%%%%%%%%%%%%%%%%%%%%%%%%%%%%%%%%%%%%%%%%%%%
%%%%%%%%%%%%%%%%%%%%%%%%%%%%%%%%%%%%%%%%%%%%%%%%%%%%%%%%%%%%%%%%%%%%
\subsection{Diffusion Expert Augmented Training}
The second enhancement to the SKIPP pipeline focuses on its training framework, which originally relied on a standard imitation learning approach based on supervised behavioral cloning (BC). The BC policy was trained to map observed states to expert actions by minimizing a pixel-wise binary cross-entropy (BCE) loss between the generated path image and the expert path image, effectively modeling the conditional probability \( p(a|s) \). This setup enabled the model to learn direct state–action correspondences from demonstrations; however, it lacked a mechanism to capture the underlying structural variability and multi-modal nature of expert behaviors beyond supervised matching.

The SPADE framework extends SKIPP by introducing diffusion model augmentation into the behavioral cloning pipeline. Instead of relying solely on supervised imitation, SPADE incorporates the diffusion model–augmented behavioral cloning (DBC) principle proposed in \cite{chen2024diffusionmodelaugmentedbehavioralcloning}. In DBC, a high-capacity diffusion model serves as an auxiliary signal that evaluates how well the BC policy’s predicted state–action pairs align with the expert distribution. This mechanism enables the transfer of structural knowledge from the diffusion model to the compact BC policy without directly imitating the diffusion model’s outputs. By modeling both the conditional probability \( p(a|s) \) and the joint probability \( p(a, s) \), DBC encourages the BC network to generalize to unseen scenarios while maintaining consistency with expert-like patterns. In this configuration, the diffusion model receives a four-channel tensor comprising the occupancy grid, start, goal, and path, and predicts all four channels jointly.

Building upon this work, we extend \cite{chen2024diffusionmodelaugmentedbehavioralcloning} by replacing the base diffusion model—which learns the joint probability distribution of state-action pairs $p(a,s)$ from the dataset—with an image-conditioned diffusion model. This modified model takes the state as input and learns the conditional probability distribution $p(a|s)$ instead. In our proposed variant, the diffusion model receives the occupancy grid, start, and goal as conditioning inputs and predicts only the path channel. Conditioning is applied through FiLM \cite{perez2017filmvisualreasoninggeneral} layers, which modulate the diffusion network’s activations using affine transformations derived from the conditioning inputs. This setup explicitly models \( p(a|s) \), providing targeted guidance to the BC network on whether a generated path is consistent with expert demonstrations given a specific environment and start–end configuration.

\begin{figure}[htbp]
    \centering
        \includegraphics[width=0.95\linewidth]{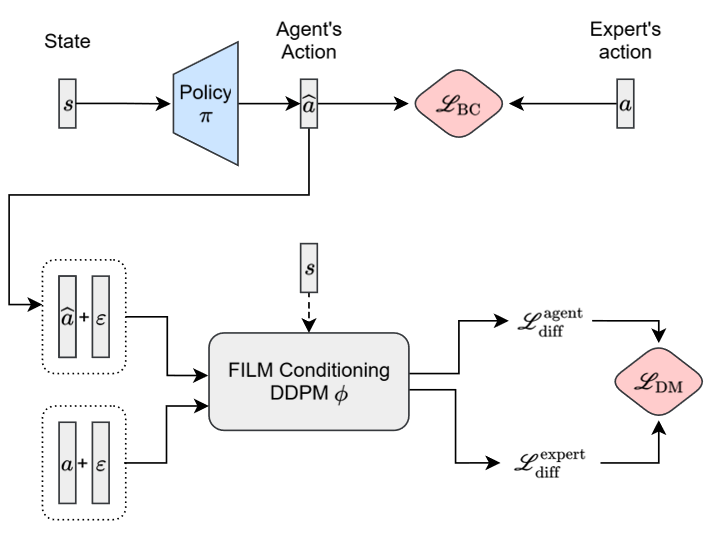}
    \caption{Diagram from \cite{chen2024diffusionmodelaugmentedbehavioralcloning} modified with the enhanced Cond-DBC pipeline with the key differences being the model $\phi$ noised predicted action and ground truth action inputs and conditioned by the state $s$. The BC model $\pi$ takes $s$ as input, predicts an action $\hat{a}$, a BC loss $\mathcal{L_{BC}}$ closes the gap between $a$ and $\hat{a}$ directly. Also, $a$ and $\hat{a}$ are noised and fed through the FiLM conditioned model $\phi$ where the margin between the loss output from both inputs is minimized through $\mathcal{L_{DM}}$.}
    \label{fig:cond-dbc_diagram}
\end{figure}

Once trained, the diffusion model estimates the likelihood that a path originates from the expert dataset by comparing reconstructed noise between generated and expert samples. During BC training, we define the diffusion margin loss \( \mathcal{L}_{DM} \) as the difference between the diffusion reconstruction loss on the generated path \( L_{\text{diff,agent}} \) and the expert path \( L_{\text{diff,expert}} \):
\begin{equation}
L_{\text{diff,agent}} = \mathcal{L}_{\text{diff}}(s, \hat{a}, \phi) = \mathbb{E}_{s \sim D, \hat{a} \sim \pi(s)} \left[ \| \hat{\epsilon}(s, \hat{a}, n) - \epsilon \|^2 \right]
\end{equation}
\begin{equation}
L_{\text{diff,expert}} = \mathcal{L}_{\text{diff}}(s, a, \phi) = \mathbb{E}_{(s,a) \sim D} \left[ \| \hat{\epsilon}(s, a, n) - \epsilon \|^2 \right]
\end{equation}
\begin{equation}
\mathcal{L}_{DM} = \mathbb{E}_{(s,a) \sim D, \hat{a} \sim \pi(s)} \left[ \max \left( L_{\text{diff,agent}} - L_{\text{diff,expert}}, 0 \right) \right]
\end{equation}
where, $\hat{\epsilon}$ is the noise predicted by the estimation model  $\phi$, $\epsilon$ is the ground truth noise at level $n$, $\pi(s)$ is the agent policy and $D$ is the dataset distribution.
The total training objective combines the BC loss with the diffusion margin loss:
\begin{equation}
\mathcal{L}_{\text{total}} = \mathcal{L}_{BC} + \lambda \mathcal{L}_{DM}
\end{equation}
where \( \lambda \) balances the influence of the diffusion guidance signal. In DBC, this loss is computed over all four channels, though the path channel dominates due to negligible reconstruction error on static inputs (occupancy, start, goal). In contrast, the FiLM-conditioned version applies the diffusion process only to the path channel, using the other inputs as conditioning variables. A diagram of the extension we propose is show in Figure \ref{fig:cond-dbc_diagram}.

SPADE preserves the efficiency of traditional behavioral cloning while injecting a generative prior through diffusion-based supervision. The diffusion model encodes multi-modal structure and long-range spatial consistency that a lightweight policy network alone would struggle to capture, leading to more stable learning and better generalization in unseen environments.
\section{Experiments and Results}
\subsection{Dataset}
\begin{figure}[htbp]
    \centering
        \includegraphics[width=0.95\linewidth]{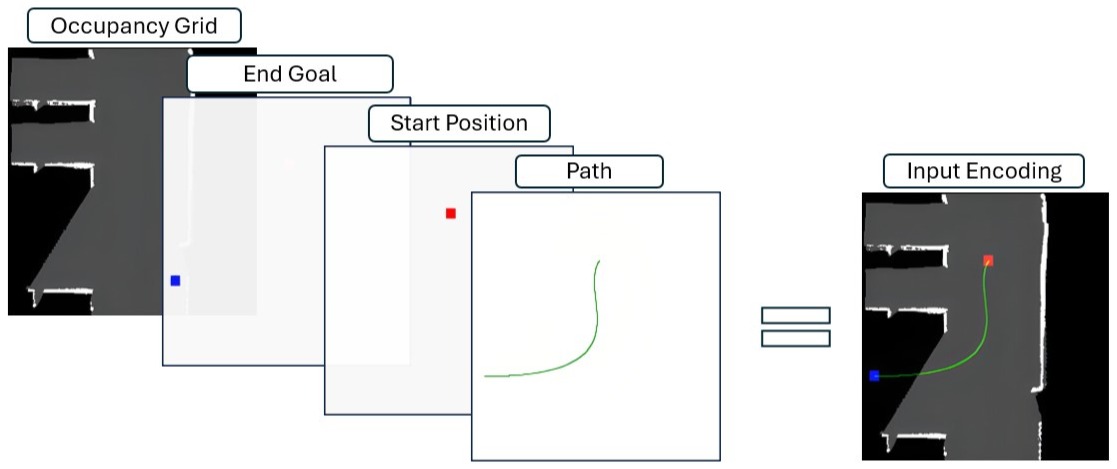}
    \caption{Occupancy grid where black pixels represent undiscovered areas, grey pixels represent navigation-allowed areas and the white pixels represent obstacles adapted from \cite{10924509}.}
    \label{fig:input_encoding}
\end{figure}

Using the newly developed annotation tool, we generated 20,000 instances divided equally between the two shapes considered in \cite{10924509}: the L-shape and the U-shape. Each instance to be loaded into our models consists of four $*(128\times128)$ binary images, as shown in Figure \ref{fig:input_encoding}. We collected data based on ten open-source occupancy grids, maximizing coverage for each map during data collection. To complete the data collection for one map, we launched last-mile delivery scenarios in different areas of the map. Each map contains multiple scenarios (typically 10), and each scenario comprises all robot states or data instances. For each shape, we considered a set of possible shape orientations, path lengths, and general variations such as small or large tails and sharper or wider turning radius. We also generated 2,000 data points as a test dataset (for both shapes) based on an industrial map. The data was then post-processed to smooth and clean it, removing any unwanted artifacts, conflicting poses, or path imperfections as per section \ref{sec:data_annotation_tool}.
\subsection{Model Setup}
In this work, we optimize model deployability and on-edge inference while simultaneously increasing performance and improving generalization. Therefore, we train a set of BC models with varying levels of numbers of trainable parameters in the SPADE framework; a small model with 117,025 parameters, a medium model with 1,926,209 parameters and a large model with 31,114,785 parameters. These parameters scale according to the breadth and the width of the encoder and decoders of the same U-Net architecture considered in \cite{10924509}. These models take as input a $(3\times128\times128)$ tensor where one channel represents the local occupancy grid and the two other channels represent the start and the end points all modeled as mentioned as binary images. The DDPM diffusion model has a ConvNext backbone \cite{liu2022convnet2020s} with progressive increase in channel dimensions from $128$ up to $2048$ having approximately 1 billion parameters. This model takes as input a $(4\times128\times128)$ tensor having the path as additional modality. The conditional diffusion model architecture was mainly inspired from \cite{chi2024diffusionpolicyvisuomotorpolicy} which integrates the FiLM conditioning layer in the encoder layers of the main encoder-decoder network. Our implementation of this model results in 9,218,561 trainable parameters. This model takes as input two main tensors; a $(1\times128\times128)$ noise matrix and the conditioning input represented as a $(3\times128\times128)$ tensor representing the occupancy grid, start and end points which all pass through the FiLM module. We train a total of three SKIPP BC and two diffusion models as baseline. Using the SPADE framework, we then train three additional BC models enhanced with an unconditional diffusion expert, referred to as DBC, and three more models augmented with a FiLM-conditioned diffusion expert, which we refer to as Cond-DBC.
\subsection{Evaluation Metrics}
To comprehensively evaluate the quality and accuracy of our generated paths, we employ three complementary metrics: Fréchet Inception Distance (FID), Absolute Pose Error (APE), and Hausdorff Distance. Each metric captures different aspects of path quality, from distributional similarity to pose-wise accuracy and worst-case deviation.
\subsubsection{Fréchet Inception Distance (FID)}
\label{sec:fid}
\begin{figure}[htbp]
    \centering
        \includegraphics[width=0.24\linewidth]{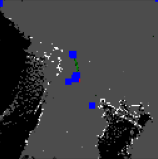}
    \hfill
    \includegraphics[width=0.24\linewidth]{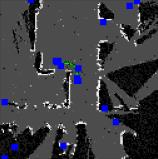}
    \hfill
    \includegraphics[width=0.24\linewidth]{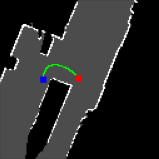}
    \hfill
    \includegraphics[width=0.24\linewidth]{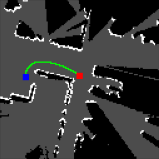}
    \caption{Comparison showing that FID effectively captures generation quality. Two examples of poor generations (images 1-2, left) with high FID scores: occupancy grid FID of 78, path FID of 72, start point FID of 60 and end point FID of 80. Two examples of good generations (images 3-4, right) with low FID scores: occupancy grid FID of 17, path FID of 8, start point FID of 1.6 and end point FID of 6.1.}
    \label{fig:fid_samples}
\end{figure}

The Fréchet Inception Distance (FID) \cite{heusel2018ganstrainedtimescaleupdate} measures the distributional similarity between generated and real path datasets. Originally developed for evaluating generative models in computer vision, FID computes the Fréchet distance between two multivariate Gaussian distributions fitted to feature representations of the real and generated data. In our context, we extract features from path representations and compute:
\begin{equation}
    \text{FID} = \|\mu_r - \mu_g\|^2 + \text{Tr}\left(\Sigma_r + \Sigma_g - 2\sqrt{\Sigma_r \Sigma_g}\right)
\end{equation}
where $\mu_r$ and $\mu_g$ are the mean feature vectors of real and generated paths respectively, $\Sigma_r$ and $\Sigma_g$ are their covariance matrices, and $\text{Tr}(\cdot)$ denotes the trace operator. Lower FID scores indicate better alignment between the generated and real path distributions, suggesting that the model captures the overall characteristics of expert demonstrations. A visual representation is shown in figure \ref{fig:fid_samples}.
\subsubsection{Absolute Pose Error (APE)}
The Absolute Pose Error (APE) \cite{grupp2017evo, 8593941} quantifies the pose-wise deviation between a generated trajectory and its corresponding ground truth expert demonstration. For a generated path $\mathbf{p}_g = \{p_g^1, p_g^2, \ldots, p_g^n\}$ and ground truth path $\mathbf{p}_r = \{p_r^1, p_r^2, \ldots, p_r^n\}$, APE is defined as:
\begin{equation}
    \text{APE} = \frac{1}{n}\sum_{i=1}^{n} \|p_g^i - p_r^i\|_2
\end{equation}
where $n$ is the number of waypoints in the path and $\|\cdot\|_2$ denotes the Euclidean distance. APE is commonly used in SLAM and trajectory evaluation to assess the absolute accuracy of estimated poses against ground truth. Lower APE values indicate paths that closely follow expert trajectories.
\begin{figure}[htbp]
    \centering
        \includegraphics[width=0.8\linewidth]{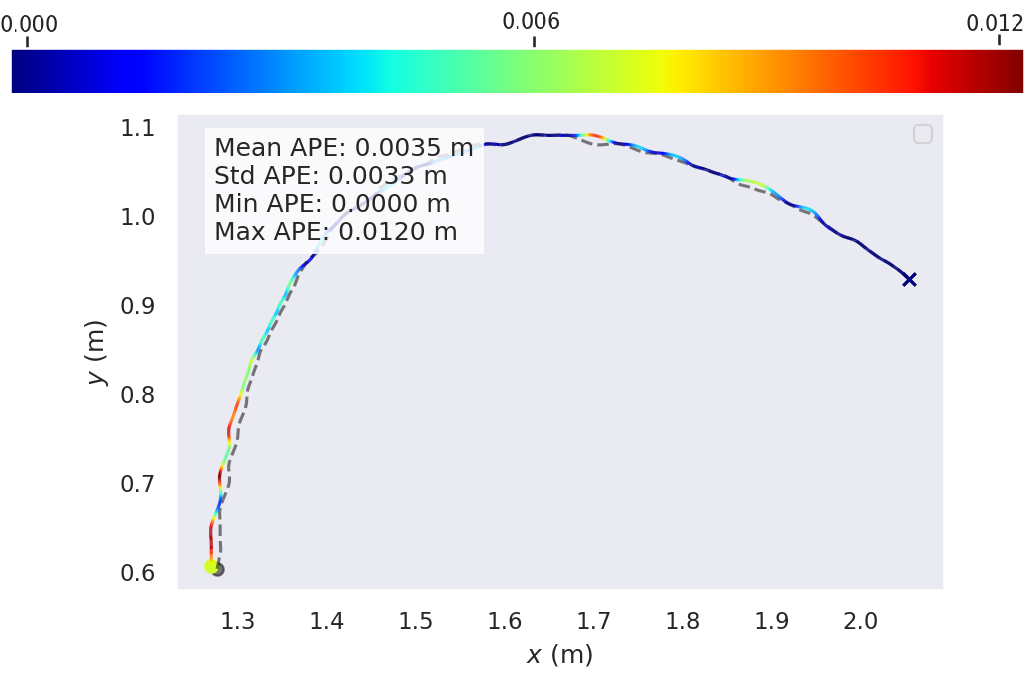}
    \caption{Color coded trajectory analysis of a sample U-shape path showing a mean APE along the path of 0.0035 meters compared to the ground truth path shown in dashed lines. Also shown are a low maximum APE of 0.012 meters and small standard deviation of 0.003 meters which demonstrates accurate and robost pose predictions.}
    \label{fig:ape_sample}
\end{figure}
\subsubsection{Hausdorff Distance}
\begin{figure}[htbp]
    \centering
        \includegraphics[width=1.0\linewidth]{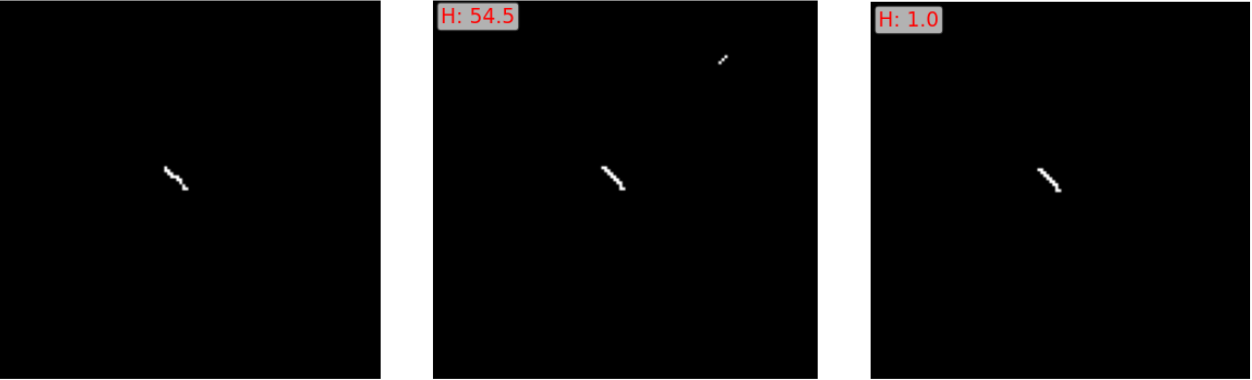}
    \caption{Examples of path predictions and their corresponding Hausdorff distances: (left) ground truth trajectory, (middle) artifact-containing prediction with a Hausdorff distance of 54.5, and (right) well-aligned prediction with a Hausdorff distance of 1.0.}
    \label{fig:haus_sample}
\end{figure}

While APE captures average deviation, the Hausdorff Distance \cite{576361} measures the maximum deviation between two paths, providing insight into worst-case performance. For two point sets $A = \{a_{1}, ..., a_{p}\}$ and $B = \{b_{1}, ..., b_{q}\}$, the Hausdorff Distance is defined as:
\begin{equation}
H(A, B) = max(h(A, B), h(B, A))
\end{equation}
Where,
\begin{equation}
h(A, B) = \max_{a \in A} \min_{b \in B} \|a - b\|
\end{equation}
and $\|\cdot\|$ is an underlying norm on the points of $A$ and $B$ and could be for example the $L_2$ or Euclidean norm. This metric identifies the largest positional discrepancy between paths, which is critical for safety-critical applications where maximum deviation bounds are important. Lower Hausdorff distances indicate better worst-case alignment. In our application, the Hausdorff distance serves as an effective indicator for detecting artifacts generated by our models. We define artifacts as either disconnected path segments or pixels that do not belong to the longest continuous path. Since we are primarily concerned with the presence of artifacts rather than their precise magnitude, we focus on the distribution of Hausdorff distances across the test set rather than individual values. By establishing an empirically determined threshold, we can identify cases where artifacts exist based on distances that exceed this limit. Model performance is evaluated by counting the number of samples with Hausdorff distances above versus below the threshold, effectively distinguishing between acceptable predictions and those containing artifacts. Through statistical analysis of a large number of predictions and qualitative assessment of artifact existence, path continuity, and overall distribution, we determined that a threshold value of 19 is well-suited for our use case. Therefore, we refer to this metric as $Hd_{19}$, an integer value representing the count of instances in our dataset with a Hausdorff distance larger than 19. Sample Hausdorff distances are shown in Figure \ref{fig:haus_sample}.
Together, these three metrics provide a comprehensive evaluation framework: FID assesses distributional fidelity, APE measures pose-wise accuracy, and Hausdorff Distance bounds worst-case errors.

\begin{table*}[htbp]
\centering
\caption{Benchmark comparison across different model sizes and training approaches for L-shape and U-shape behaviors}
\label{tab:benchmark}
\small
\begin{tabular}{llccccccccc}
\hline
\textbf{Size} & \textbf{Approach} & \multicolumn{3}{c}{\textbf{L-Shape}} & & \multicolumn{3}{c}{\textbf{U-Shape}} \\
\cline{3-5} \cline{7-9}
& & FID ↓ & APE (cm) ↓ & $Hd_{19}$ ↓ & & FID ↓ & APE (cm) ↓ & $Hd_{19}$ ↓ \\
\hline

\multirow{3}{*}{Small} 
& BC (Baseline) & 42.27 & 4.43 & \textbf{28} & & 35.51 & 6.79 & \textbf{52} \\
& DBC (Uncond. Diff.) & 24.69 & \textbf{2.92} & 62 & & 29.86 & \textbf{6.55} & 124 \\
& Cond-DBC (Cond. Diff.) & \textbf{23.87} & 3.14 & 43 & & \textbf{27.65} & 6.72 & 52 \\
\hline

\multirow{3}{*}{Medium} 
& BC (Baseline) & 31.25 & 3.48 & 6 & & 27.80 & 6.01 & 11 \\
& DBC (Uncond. Diff.) & 16.51 & 2.21 & \textbf{1} & & 15.01 & 4.51 & \textbf{7} \\
& Cond-DBC (Cond. Diff.) & \textbf{13.58} & \textbf{1.46} & 2 & & \textbf{13.97} & \textbf{4.01} & 9 \\
\hline

\multirow{3}{*}{Large} 
& BC (Baseline) & 20.45 & 2.41 & 5 & & 17.12 & 5.20 & 7 \\
& DBC (Uncond. Diff.) & 15.72 & 1.89 & \textbf{0} & & \textbf{9.59} & \textbf{2.87} & 10 \\
& Cond-DBC (Cond. Diff.) & \textbf{12.47} & \textbf{1.47} & 5 & & 12.37 & 3.73 & \textbf{0} \\
\hline
\end{tabular}
\end{table*}

\subsection{Benchmark}
To comprehensively evaluate our framework across different model scales, we conducted a benchmark comparison of three approaches: standard behavioral cloning (BC), BC augmented with conditional diffusion or Cond-DBC, and BC augmented with unconditional diffusion or DBC. Among the standalone diffusion models, the unconditional DDPM achieved one of the best overall outcomes, producing the lowest FID for the path (2.43 for the U-shape and 4.84 for the L-shape) with an inference time of approximately 45{,}000~ms (45~s). The conditional DDPM obtained slightly higher FID values ( 6.75 for the U-shape and 13.43 for the L-shape) while achieving the lowest APE values (2.74~cm for the U-shape and 1.42~cm for the L-shape) and demonstrating a faster inference time of around 18{,}000~ms (18~s). Despite their strong generative performance, both diffusion variants remain unsuitable for real-time deployment due to their high inference latency.

Each augmentation approach was tested with small, medium, and large BC model architectures. Table~\ref{tab:benchmark} presents the results across all configurations. The results demonstrate that introducing a diffusion model, whether unconditioned or conditioned, led to substantial improvements across all architectures. The diffusion-based policies consistently achieved lower FID and APE values compared to the baseline SKIPP BC, which underperformed in both metrics. 

Between the two diffusion-augmented strategies, Cond-DBC generally provided the strongest results, outperforming DBC in nearly all cases. The only exception occurred with the small model size, where DBC achieved a marginally better APE (2.92~cm compared to 3.14~cm, respectively), while Cond-DBC achieved the best FID (23.87 compared to 24.69). When using a larger backbone for the L-shape, Cond-DBC effectively closed the gap in both FID and APE relative to the standalone conditional DDPM, showing no statistically significant difference on either seen or unseen maps.

The $Hd_{19}$ results reveal an interesting trend across model scales and training approaches. While baseline BC models show relatively low artifact counts for medium and large architectures (6--7 instances), the small BC model struggles significantly, with 28--52 instances for L-shape and U-shape respectively, indicating poor generalization at limited capacity. Interestingly, augmenting small models with unconditional diffusion (DBC) paradoxically increases artifact generation (62 and 124 instances), suggesting that the unconditioned diffusion guidance may introduce spurious paths when the student network lacks sufficient capacity to properly interpret the joint distribution signal. In contrast, Cond-DBC maintains artifact counts comparable to or better than baseline BC even for small models, demonstrating that image-conditioned guidance provides more focused and interpretable signals that small networks can effectively leverage. For medium and large models, both DBC and Cond-DBC achieve near-zero artifact rates (0--10 instances), with DBC occasionally outperforming Cond-DBC, indicating that sufficient model capacity enables effective learning from both guidance types. This analysis underscores the importance of matching the diffusion guidance strategy to the student model capacity for optimal artifact-free generation.
    
Moreover, diffusion-augmented policies offer significant efficiency advantages. A medium Cond-DBC model can be favored for deployment over the large baseline model while using approximately 93.8\% fewer trainable parameters. For instance, the medium-size Cond-DBC achieved 39.1\% lower APE, 33.5\% lower FID, and 60\% lower $Hd_{19}$ than the large BC baseline. The same trend is observed for DBC, although Cond-DBC consistently maintains a slight performance advantage across all scales.

These results have significant implications for practical AMR deployment. By achieving diffusion-level performance with compact models, SPADE enables on-robot inference without requiring cloud computation or specialized hardware accelerators. The medium Cond-DBC model, with only 1.9 million parameters, achieves performance comparable to billion-parameter diffusion models while maintaining real-time inference capabilities suitable for embedded systems. This orders-of-magnitude reduction in computational requirements, combined with improved generalization—makes learning-based, preference-aware path planning viable for resource-constrained industrial robots. The framework thus bridges the gap between the representational power of large generative models and the strict latency, memory, and power constraints of real-world robotic systems, enabling deployable solutions that adapt to operator preferences without sacrificing performance or requiring expensive hardware upgrades.
\section{Ablation Studies}
To evaluate the effectiveness of our framework in adapting to new behaviors and environments, we conducted a comprehensive ablation study focused on transfer learning from the L-shape to the U-shape behavior on a new industrial map. We generated two new datasets for the U-shape behavior, each containing 1,000 instances for training and testing respectively. Our ablation study examines two key aspects: (1) the benefit of fine-tuning our augmented BC over the standard SKIPP model, and (2) the additional benefit of incorporating diffusion-based expert guidance (trained on L-shape data) during fine-tuning, comparing models fine-tuned with and without this cross-behavior knowledge transfer on the new behavior and map.

We compare three fine-tuning approaches, all starting from models pre-trained on L-shape behavior. The first baseline fine-tunes the standard SKIPP model on the new U-shape data. The second approach fine-tunes our augmented BC model without the SPADE pipeline (standard fine-tuning). The third approach represents our complete framework: fine-tuning the augmented BC model using the SPADE pipeline with a pre-trained L-shape conditional diffusion expert. As shown in Table \ref{tab:ablation}, our complete framework achieves the best performance across all metrics. Compared to fine-tuning the standard SKIPP model, our complete approach improves FID score by 37.1\%, APE by 23.1\%, and $Hd_{19}$ by 62\%. To isolate the contribution of the SPADE pipeline, we compare fine-tuning with and without it on the augmented BC model, which reveals that diffusion-based expert guidance provides an additional 6.2\% improvement in FID, 3.72\% in APE, and 26.92\% in $Hd_{19}$. These results validate that each component of our framework contributes meaningfully to cross-behavior transfer, with the complete SPADE pipeline achieving the strongest generalization to new behaviors and environments.
\begin{table}[htbp]
\centering
\caption{Ablation study results for cross-behavior transfer from L-shape to U-shape via fine-tuning}
\label{tab:ablation}
\begin{tabular}{lcccc}
\hline
Fine-tuning Approach & FID ↓ & APE (cm) ↓ & $Hd_{19}$ ↓ \\
\hline
SKIPP BC baseline & 37.43 & 7.75 & 50 \\
Augmented BC only & 25.10 & 6.19 & 26 \\
Augmented BC+ Diffusion & \textbf{23.55} & \textbf{5.96} & \textbf{19} \\
\hline
\end{tabular}
\end{table}

To further validate our framework's effectiveness, we conducted an additional experiment comparing training from scratch on the U-shape behavior. We trained two models: a standard SKIPP model and an augmented BC model with a pre-trained diffusion expert during training. As shown in Table \ref{tab:ablation_scratch}, training with our complete SPADE pipeline achieved FID score of 44.14, APE of 8.26~cm, and $Hd_{19}$ of 7. Compared to training the standard SKIPP model from scratch, our SPADE pipeline improved FID by 4.21\%, and $Hd_{19}$ by 63.16\%. These results validate that each component of our framework contributes meaningfully to both transfer learning and training from scratch scenarios, with the complete SPADE pipeline achieving the strongest generalization to new behaviors and environments.

\begin{table}[htbp]
\centering
\caption{Ablation study results for training from scratch on U-shape behavior}
\label{tab:ablation_scratch}
\begin{tabular}{lcccc}
\hline
Training Approach & FID ↓ & APE (cm) ↓ & $Hd_{19}$ ↓ \\
\hline
SKIPP BC & 46.08 & \textbf{8.15} & 19 \\
Augmented BC + & \textbf{44.14} & 8.26 & \textbf{7} \\
pre-trained diff & & & \\
\hline
\end{tabular}
\end{table}

\section{Conclusion}
This work addresses critical limitations in sketch-guided path planning for Autonomous Mobile Robots by introducing SPADE, a framework that combines diffusion-based expert guidance with lightweight behavioral cloning models. We presented two key contributions: a robust, open-source annotation tool built on ROS 2 that ensures reproducibility and extensibility, and a novel training strategy that leverages both unconditional and image-conditioned diffusion models to enhance generalization while maintaining deployment efficiency.

Our experimental results demonstrate that SPADE significantly outperforms the baseline SKIPP approach, achieving 39.1\% lower APE and 33.5\% lower FID while using 93.8\% fewer trainable parameters. Notably, our medium-sized Cond-DBC model achieves performance comparable to standalone diffusion models that are orders of magnitude larger and require significantly longer inference times (18-45 seconds vs. real-time execution). This substantial reduction in model complexity without sacrificing performance represents a crucial advancement for practical AMR deployment, where computational constraints and real-time requirements are paramount.

The ablation studies further validate the robustness of our framework, showing effective transfer learning capabilities across different behaviors and environments. When fine-tuning from L-shape to U-shape behavior on a new industrial map, our complete SPADE pipeline achieved 37.1\% improvement in FID and 62\% reduction in artifact count compared to baseline fine-tuning. These results demonstrate that diffusion-based expert guidance enables small, efficient models to achieve high generalization capabilities—a critical requirement for real-world robotic systems where model size, inference speed, and adaptability must be carefully balanced.

Despite these advances, several promising directions remain for future work. First, extending the framework to handle dynamic environments with moving obstacles would significantly broaden its applicability to real-world scenarios where pedestrians, vehicles, or other robots share the workspace. Second, developing more sophisticated training strategies for diffusion experts could provide richer behavioral priors, enabling more effective fine-tuning on diverse behaviors and map topologies with minimal additional data. Finally, investigating multi-modal diffusion conditioning that incorporates temporal information or higher-level task specifications could further enhance the framework's flexibility and practical utility.

SPADE demonstrates that lightweight, deployable models can achieve strong generalization when guided by high-capacity experts during training, offering a practical path forward for learning-based path planning in resource-constrained robotic systems.

\bibliographystyle{ACM-Reference-Format} 
\bibliography{sample}

%%%%%%%%%%%%%%%%%%%%%%%%%%%%%%%%%%%%%%%%%%%%%%%%%%%%%%%%%%%%%%%%%%%%%%%%

\end{document}